\documentclass[authorversion,nonacm]{acmart}

\usepackage{url}

\setcopyright{none}

\graphicspath{{./pics/}{./}}

\usepackage[absolute]{textpos}
\begin{document}

\begin{textblock}{151}(26,255)
\noindent \scriptsize
\copyright\ 2021 The Authors. 
This is the authors' accepted manuscript version of the work. It is posted here for your personal use. Not for redistribution. The definitive version was published as: Joni Korpihalkola, Tuomo Sipola, Samir Puuska, and Tero Kokkonen. 2021. One-Pixel Attack Deceives Computer-Assisted Diagnosis of Cancer. In \emph{2021 4th International Conference on Signal Processing and Machine Learning (SPML 2021), August 18–20, 2021, Beijing, China}. ACM, New York, NY, USA, 7 pages. https://doi.org/10.1145/3483207.3483224
\end{textblock}

\title{One-Pixel Attack Deceives Computer-Assisted Diagnosis of Cancer}


\author{Joni Korpihalkola}
\email{joni.korpihalkola@jamk.fi}
\orcid{0000-0001-6434-1240}
\affiliation{
  \institution{Institute of Information Technology, JAMK University of Applied Sciences}
  \city{Jyv{\"a}skyl{\"a}}
  \country{Finland}
}

\author{Tuomo Sipola}
\email{tuomo.sipola@jamk.fi}
\orcid{0000-0002-2354-0400}
\affiliation{
  \institution{Institute of Information Technology, JAMK University of Applied Sciences}
  \city{Jyv{\"a}skyl{\"a}}
  \country{Finland}
}

\author{Samir Puuska}
\email{sapepuus@student.jyu.fi}
\orcid{0000-0003-0785-575X}
\affiliation{
  \institution{Faculty of Information Technology, University of Jyv{\"a}skyl{\"a}}
  \city{Jyv{\"a}skyl{\"a}}
  \country{Finland}
}

\author{Tero Kokkonen}
\email{tero.kokkonen@jamk.fi}
\orcid{0000-0001-9988-6259}
\affiliation{
  \institution{Institute of Information Technology, JAMK University of Applied Sciences}
  \city{Jyv{\"a}skyl{\"a}}
  \country{Finland}
}

\renewcommand{\shortauthors}{Korpihalkola, et al.}

\begin{abstract}
Computer vision and machine learning can be used to automate various tasks in cancer diagnostic and detection. 
If an attacker can manipulate the automated processing, the results can be devastating and in the worst case lead to wrong diagnosis and treatment. 
In this research, the goal is to demonstrate the use of one-pixel attacks in a real-life scenario with a real pathology dataset, TUPAC16, which consists of digitized whole-slide images.
We attack against the IBM CODAIT's MAX breast cancer detector using adversarial images. 
These adversarial examples are found using differential evolution to perform the one-pixel modification to the images in the dataset. 
The results indicate that a minor one-pixel modification of a whole slide image under analysis can affect the diagnosis by reversing the automatic diagnosis result. 
The attack poses a threat from the cyber security perspective: the one-pixel method can be used as an attack vector by a motivated attacker.
\end{abstract}



\maketitle

\section{Introduction}

\subsection{Cancer detection as a target}

Cancer is one of the most common causes of death in the western world. 
The number of detected cancers of a determined type in a defined population during a year is expressed as cancer
incidence rate (CIR), commonly formed as the number of cancers per 100,000 population~\cite{NIH-1}. 
According to the U.S. National Cancer Institute at the National Institutes of Health, the CIR, based on
2013--2017 statistics in the U.S., is 442.4 per 100,000 men and women per year~\cite{NIH-2}. 
In many common types of cancers, early
detection is a key factor in improving the prognosis~\cite{diest675}. 
However, early detection is a time-consuming activity. 
Automating some of the work with techniques such as machine learning and computer vision 
can lead to faster detection, increased throughput, and reduced costs~\cite{zhang2017network,nasief2019machine}, e.g., when a count of cells needs to be made from an image \cite{veta2016mitosis}. 
Although various approaches for analyzing shapes have been proposed in the literature,
one of the newer approaches is to use an artificial neural
network for detecting the desired properties~\cite{khosravi2018deep,bera2019artificial,alom2019advanced}.

From the cyber security standpoint this increased automation means
increased attack surface. 
Although cyber
operations against computers, networks, and data required
for administering medical care are prohibited under
international law, there has been a steady increase in
attacks against them~\cite{schmitt2017}. Awareness has an important role in cyber
security of the healthcare sector, as stated by Rajam{\"a}ki et al.~\cite{Rajamaki2018}: \textit{``The
highest concern for healthcare organizations is the employee negligence followed
by the fear of a cyber-attack.''}
In their study, Spanakis et al.\ analyzed cyber
security in the healthcare domain and stated the fact that growth of technology utilized in
healthcare concurrently increases the attack surface and thus the risk of cyber incidents increases~\cite{Spanakis2020}. 
One possible motivation can be the capability of claiming ransoms. Modifying the automated diagnosis capability with a cyber attack may affect the treatment and in the worst case scenario lead to loss of human lives. That also raises the possibility of targeted attacks against a particular person. In conclusion, such attacks may lead to global lack of trust in automated diagnosis systems~\cite{sipola2020model}.

\subsection{Adversarial attacks}

Goodfellow et al.\ define adversarial examples as samples where an adversary
makes a small but well-chosen perturbation into an input sample causing the
artificial neural network classifier to misclassify that sample with high confidence~\cite{goodfellow2015explaining}.
If an adversary possesses a fast way of generating the adversarial examples, they can be used
to mount various types of attacks against systems that utilize neural networks
for classification tasks.

Papernot et al.\ state that models of machine learning are vulnerable to modified (malicious) inputs and on that account they introduced a black-box attack against deep neural networks without knowledge of the classifier training data or model~\cite{Papernot_2017}. Such attack methods have been introduced in some real world scenarios, for example, Stokes et al.~\cite{Stokes_2018} studied the attack, and furthermore, defence of malware detection models and image modifications against artificial intelligence (AI) based computer vision capabilities has been researched in~\cite{Kang_2020}. When attacking against computer vision and image based machine learning, pixel modification is an evident possibility. Lin et al.\ tested adversarial attacks by modifying critical pixels of the image with limitations for the number of modified pixels. They changed as few as five pixels in the general CIFAR-10 dataset using a gradient-based dual iterative fusion method~\cite{Lin_2020}. A short survey of model fooling attacks in the medical domain by Sipola et al.\ shows that at least adversarial images and patches have been used in experiments~\cite{sipola2020model}.
One-pixel attack is a more advanced method, in which only one pixel of an image is modified in order to fool the classifier~\cite{Su_2019}. Additionally, mitigation capabilities have been developed, Paul et al.~\cite{Paul_2020} introduce mitigation of adversarial attacks on medical image systems with the conclusion that its effectiveness can be decreased by adding adversarial images in the training set. In addition to this kind of robust optimization, Xu et al.~\cite{xu2020adversarial} mention the possibility of gradient masking and attack detection before forwarding the images to the actual classifier. However, the detection of attacks is beyond the scope of our study. 

In this article we show that creating adversarial examples in the context of medical
imaging is both feasible and fast. Furthermore, we show that a particular type
of one-pixel perturbation is sufficient to alter legitimate input
images in a fashion that causes the classifier to misclassify them with high
confidence.
This work is the implementation of the conceptual attack framework of using one-pixel attacks against medical imaging~\cite{Sipola_2021}.
In addition, we explore how
to create fake images that appear authentic also to
the human observer. Thirdly, we present a method for altering
existing images to achieve the goal.
We opted to use full slide microscopy images, as they
are a major target for automated analysis and digital pathology research, as well as
being readily available for scientific use.
Although the proposed methods are presented in
a cyber attack context, the results can be used to improve
or assess classifiers outside this context.

\section{Methods and experimental setup}



Two attacks were performed: mitosis-to-normal, where the objective is to minimize the confidence score value,
and normal-to-mitosis, where the objective is to maximize the confidence score value.
The former attack type alters an image containing abnormal mitosis into one where the
classifier fails to detect this with high confidence. The latter converts an image with
normal mitosis into one where the classifier misclassifies it being an abnormal mitosis.

\subsection{Attack target}

The dataset used in this research is from the Tumor Proliferation Assessment Challenge 2016 (TUPAC16)~\cite{tupac16, tupacpaper}. The dataset consists of 500 whole slide light microscopy images with known tumor proliferation scores, ground truth labels for the training set, as well as region of interest location data for 148 images.
The dataset was preprocessed by a script from IBM CODAIT Center for Open-source Data \& AI Technologies' \emph{deep-histopath} repository,\footnote{Available online at: \url{https://github.com/CODAIT/deep-histopath}} which split the whole slide image into 64-by-64 pixel PNG-format images. The images were marked either `mitosis' or `normal' according to the provided labeling. 

The chosen target classifier was IBM CODAIT's MAX breast cancer detector, which is here used as a pretrained black-box neural network classifier~\cite{dusenberry2018}. This classifier was chosen for its high ranking in the TUPAC16 challenge, and the open source nature of the code. Due to the nature of artificial neural network -based classifiers, this attack method is likely to work on other TUPAC16 contest entries as well. The obtained results do not suggest a particular failure or error in the IBM CODAIT's work or approach.

To simulate a black-box attack situation, the artificial neural network is queried through a HTTP API. Only the input image and the confidence score of the artificial neural network model for the input image are known. Inference on the model was performed by converting the image to a byte string and querying the model API residing in a Docker container. The response from the API returned a confidence score for the image. The images were also filtered based on the confidence score provided by the artificial neural network. A `mitosis' labeled image with confidence score below 0.9 and `normal' labeled image with score above 0.1 are filtered out of the experiment. This way the attacks focus on the unambiguous cases that should be classified correctly by the artificial neural network. Computation time was capped at five days. Consequently, 5,343 `mitosis' and 80,725 `normal' labeled images are tested using this method. 

\subsection{Attack outline}

The goal of the attack is to find a method capable of perturbing legitimate input images in a way that causes the classifier to misclassify them with high 
confidence. It is usually in the interest of the attacker to find a perturbation that alters the original image as little as possible. The so-called \emph{one-pixel attack}
is achieved when the perturbation that causes a misclassification consists of altering just one pixel in the input image. To a human observer the difference between the original and altered image might be indistinguishable. As stated, two attacks are performed: mitosis-to-normal, where the objective is to minimize the confidence score value and normal-to-mitosis, where the objective is to maximize the confidence score value.

To carry out a black-box attack the adversary needs to make perturbations to the original image, and observe how the classifier under attack reacts. Su et al.\ proposed
a method capable of creating one-pixel perturbations using differential evolution~\cite{su2019one}. Differential evolution is an optimization method~\cite{feoktistov2006differential, price2013differential} which can be leveraged for iteratively refining the chosen perturbations until the attacker achieves the desired misclassification confidence.
In this study, we used the implementation of differential evolution in the SciPy library~\cite{2020SciPy-NMeth}.

A color digital image can be presented as a grid of pixels, where each pixel is a mix of red, green, and blue colors, corresponding to the color sensing cells in human eye.
A one-pixel perturbation can be represented by a vector:~$\mathbf{x}~=~(x,~y,~r,~g,~b)$, where $x$ and $y$ are the pixel coordinates and $r$, $g$, $b$ are the red, green and blue values of the color. All these variables are integers. The bounds for coordinates are [0,63] and the bounds for color values are [0,255].

The initial population consists of 200 one-pixel perturbation attack vectors, the vector values are initialized using Latin hypercube sampling, which ensures that each coordinate and color value is uniformly sampled inside its bounds. A larger initial population was found to increase attack success only in some rare cases, while it slowed down attack vector search considerably due to higher computation costs. The mutation factor was set at 0.5 and the recombination factor at 0.7. Larger mutation factor and lower recombination factor values were not found to impact the attack success rate in neither mitosis-to-normal nor normal-to-mitosis attacks. Maximum iterations for the evolution were set at 100, although in practice the evolution converged on average at 44 iterations in mitosis-to-normal and 39 on normal-to-mitosis attacks. 
After the initial population is created, the members of the population are iterated over. The strategy for creating trial vectors was chosen as `best1bin'.

\subsection{Attack success metric}

The first criterion for a successful attack is the number of steps in its evolution progress. Attacks that converged after the iteration of the initial population were found to not alter the confidence score at all or by very little margin. Thus, a successful attack needs to iterate the population more than once. On the other hand, the less iterations a success uses, the less computing power is used to find it. 

The second criterion is the confidence score threshold used to determine if an attack is a success. The closer the model's confidence score is to 1, the more sure the model is that the image should be labeled `mitosis' and the closer the score is to 0, the image is to be labeled `normal'. To define attacks as successful, mitosis and normal attacks should reach at least 0.5 score threshold, reducing the neural network's prediction into a coin flip. If a mitosis-to-normal attack manages to lower confidence score to 0.05 or a normal-to-mitosis attack the score to 0.95, the model is fooled to predict the opposite label with high certainty.

\section{Results}

The one-pixel attack was performed on 5,343 `mitosis' labeled images and 80,725 `normal' labeled images. The attack results were documented in a comma-separated values (CSV) file, including the name of the images used in the experiment, differential evolution parameters, original confidence score and the score after the attack. The confidence score indicates how confident the artificial neural network is that the image contains mitosis activity. The score varies between $[0,1]$, where $0$ means that the image is considered normal and $1$ means that it is considered to contain mitoses. 

\subsection{Failures due to early convergence}

Attacks where evolution converged immediately after the initial population can be considered as failed attacks. 
In mitosis-to-normal attacks, in 1,594 or approximately 30\% of the attacks the algorithm converged already after the calculation of initial population function values, while in normal-to-mitosis attacks, 80,520 or 99.7\% converged after the initial population. This is due to the tolerance value set at 0.01 and the standard deviation of the initial population being too low compared to tolerance value multiplied by mean of initial population. Lowering of the tolerance value had no impact on finding more successful populations. The tolerance value and the convergence check cause the amount of `normal' labeled images processed to be higher than `mitosis' labeled, because more evolution steps were performed on `mitosis' labeled images. If the evolution converges after the initial population values, in mitosis-to-normal attacks the attack yields only 0.06 change in confidence score on mean and in normal-to-mitosis attacks the confidence score change is 0.001 on mean.  

\subsection{Confidence scores}

The changes in the confidence score were noticeable in both attack types.
On mitosis-to-normal attack, 3,407 attacks (91\%) out of 3,749 managed to lower the artificial neural network's confidence score below 0.5 and 895 attacks (24\%) lowered it below 0.05. On normal-to-mitosis attacks, neural network's confidence score was raised higher than 0.5 on 173 out of 205 attacks (84\%) but none of the attacks managed to cross above the 0.95 score threshold. 

When looking at attacks where the differential evolution algorithm did not converge on the initial population, the median confidence score difference between the original score and score after attack reaches 0.81. 
Applying the same filter as in mitosis-to-normal attacks, the median confidence score difference in normal-to-mitosis attacks between original images and adversarial images reaches 0.27. 

\begin{figure}
\centering
\includegraphics[width=0.5\textwidth]{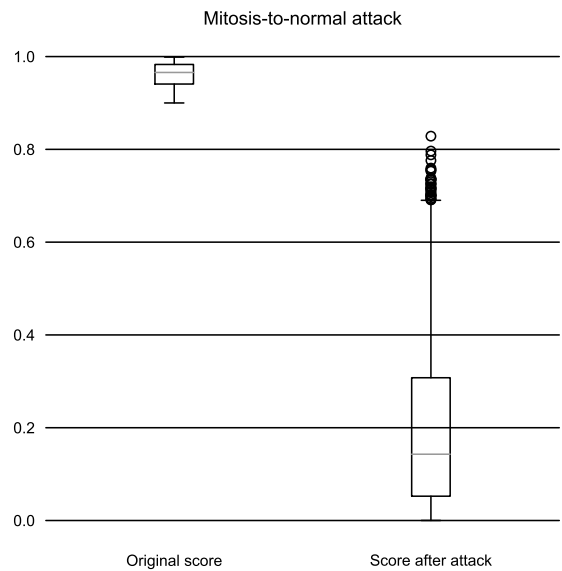}
\caption{Box plot visualization of mitosis-to-normal attack experiment confidence scores. The majority of the attacks were successful, lowering the confidence score below $0.5$ in 3,407 or approximately 91\% of the attacks. 895 or approximately 24\% of the attacks manage to lower the confidence score below $0.05$.}
\label{fig:mitosistonormalboxplot}
\end{figure}

Mitosis-to-normal attacks were successful in finding adversarial examples. Figure \ref{fig:mitosistonormalboxplot} has its center line at the median value, its box limits extending from 25\% to 75\%, its whiskers from the edges of the box to no more than 1.5 times interquartile range, ending at the farthest point in the interval and its outliers plotted as dots. The figure shows how the neural network's confidence score before the attack is on average 0.96, the maximum score is 0.99 and minimum is 0.90. After attacking the images and finding adversarial images, artificial neural network's confidence score median values are 0.1, they also reach a minimum of 0.0001 and a maximum of 0.83. The standard deviation for the scores is 0.18 and the mean is 0.20. This information is also conveyed in Table \ref{tab:mitosis-to-normal}. 

\begin{table}
\caption{Confidence score statistics for mitosis-to-normal attack, where the number of attacks is 3,749.}
\label{tab:mitosis-to-normal}
\begin{tabular}{l c c}
\toprule
                    & Before attack & After attack \\
\midrule
Maximum             & 0.99          & 0.83 \\
Mean                & 0.96          & 0.20 \\
Median              & 0.96          & 0.14 \\
Standard deviation  & 0.02          & 0.18 \\ 
Minimum             & 0.90          & 0.00011 \\
\bottomrule
\end{tabular}
\end{table}

\begin{figure}
\centering
\includegraphics[width=0.5\textwidth]{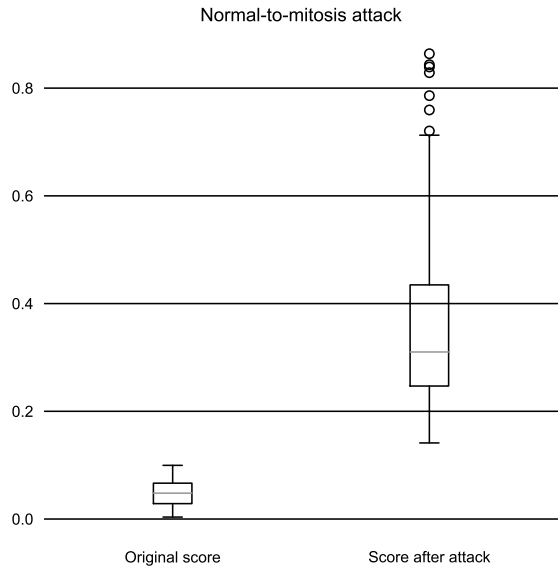}
\caption{Box plot visualization of normal-to-mitosis attack experiment confidence scores, where 173 or approximately 84\% of attacks manage to raise the artificial neural network's confidence score above $0.5$. None of the attacks manage to cross above the $0.95$ score threshold.}
\label{fig:normaltomitosisboxplot}
\end{figure}

Normal-to-mitosis attacks were also successful. Before the images are attacked, neural network's confidence scores are on average 0.048, where the minimum is 0.0036 and maximum is 0.099. Figure \ref{fig:normaltomitosisboxplot} shows this as box plot, which shares its statistical characteristics with \ref{fig:mitosistonormalboxplot}. After attacking the images, neural network's confidence score median is 0.31, the scores minimum reaches 0.14 and maximum 0.86. The standard deviation for the scores is 0.15 and the mean is 0.36. This information is also conveyed in Table \ref{tab:normal-to-mitosis}. 

\begin{table}
\centering
\caption{Confidence score statistics for normal-to-mitosis attack, where the number of attacks is 205.}
\label{tab:normal-to-mitosis}
\begin{tabular}{l c c}
\toprule
                    & Before attack & After attack \\
\midrule
Maximum             & 0.099         & 0.86 \\
Mean                & 0.048         & 0.36 \\
Median              & 0.048         & 0.31 \\
Standard deviation  & 0.025         & 0.15 \\ 
Minimum             & 0.0036        & 0.14 \\
\bottomrule
\end{tabular}
\end{table}

\subsection{Adversarial examples}

\begin{figure}
\centering
\includegraphics[width=0.7\textwidth]{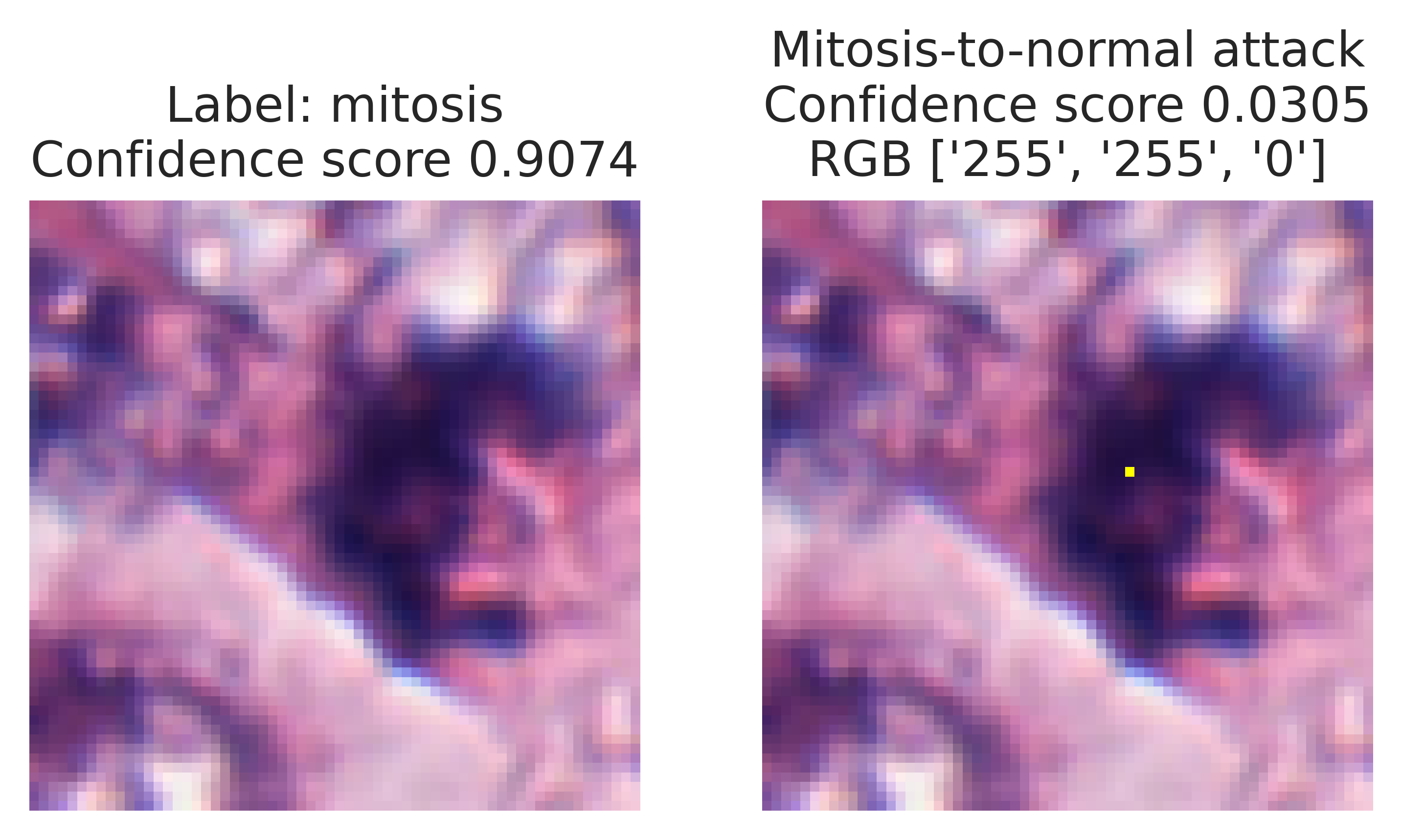}
\caption{An adversarial example that is misclassified as normal even though in reality the source image is labeled as having mitosis activity. Notice the bright yellow pixel inside the dark area in the middle right part of the image.}
\label{fig:mitosistonormal}
\end{figure}

\begin{figure}
\centering
\includegraphics[width=0.7\textwidth]{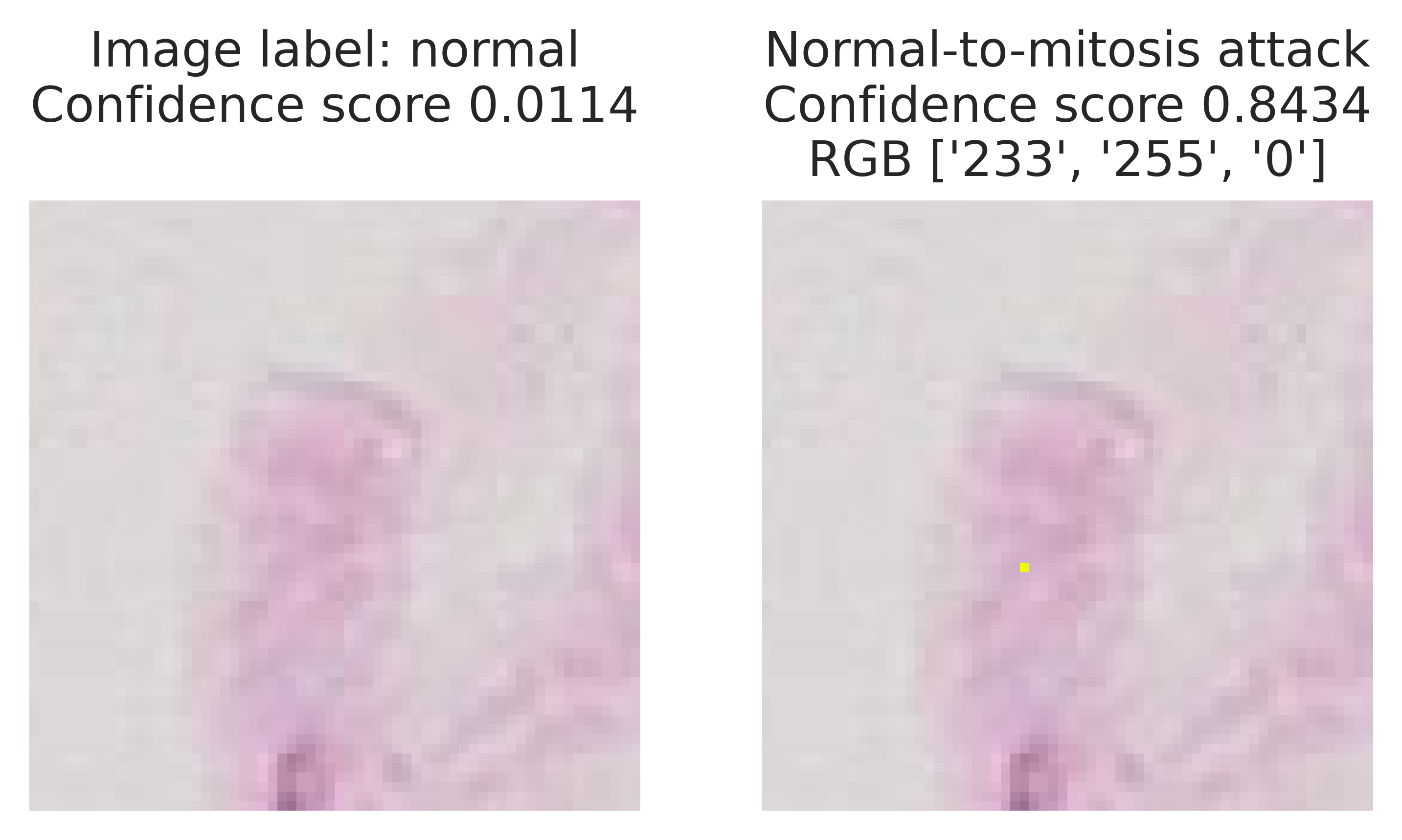}
\caption{An adversarial example that is misclassified as mitosis even though in reality the source image is labeled as having no mitosis activity. Notice the yellow-lime dot in the middle of the image.}
\label{fig:normaltomitosis}
\end{figure}

As an example, we showcase two successful one-pixel attacks. Firstly, Figure \ref{fig:mitosistonormal} shows this adversarial example that deceives the predictor to think that an image containing mitosis is a normal picture without any signs of disease. In the attacks, the most common pixel color was pure yellow, meaning RGB values (255, 255, 0), which was used in 2,214 attacks. In 122 attacks the pixel color was pure white, meaning  RGB values (255, 255, 255), which was used in 122 attacks. In the rest of the attacks the pixel colors were yellow with a slightly higher blue value. Secondly, Figure \ref{fig:normaltomitosis} shows an adversarial example that deceives the predictor to think that a picture of normal cell activity contains mitosis. The most common pixel color RGB values was pure yellow (255, 255, 0) and the second most common was pure white (255, 255, 255) and the third was pure black (0, 0, 0). There is a larger variety of colors in attack vectors than in mitosis-to-normal-attacks, but this is most likely explained due to the low amount of successful attacks. 

\begin{figure}
\centering
\includegraphics[width=0.5\textwidth]{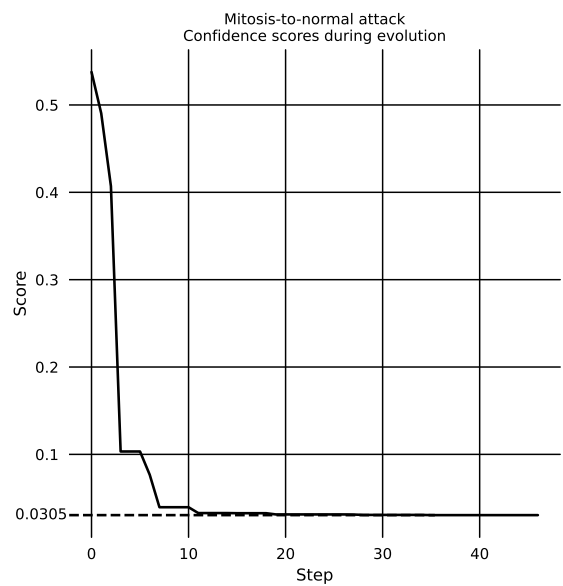}
\caption{Lowest neural network confidence scores during steps of differential evolution. Example of of an attack against one image, in this case the same as in Figure \ref{fig:mitosistonormal}.}
\label{fig:mitosistonormalgraph}
\end{figure}

\begin{figure}
\centering
\includegraphics[width=0.5\textwidth]{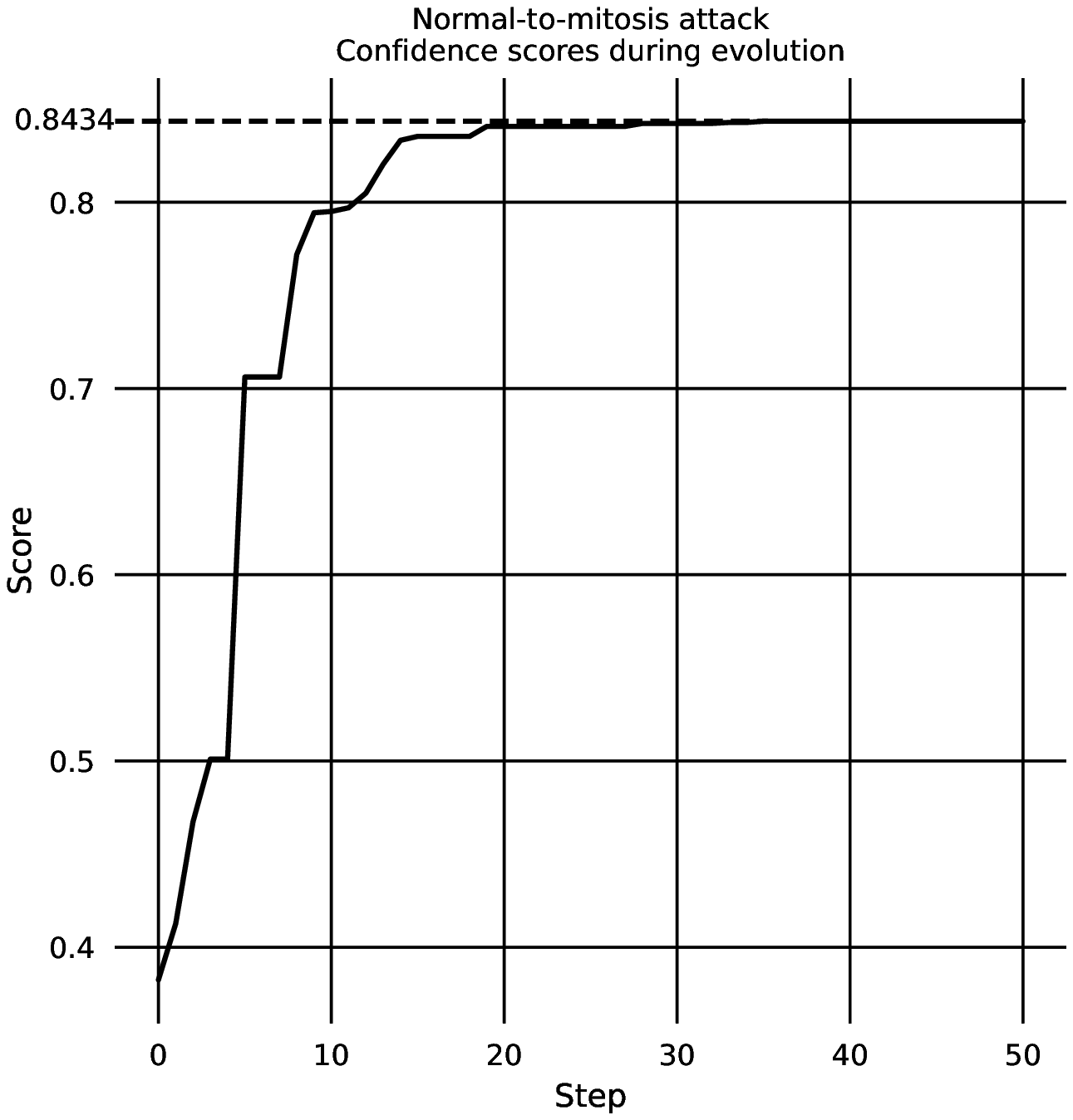}
\caption{Highest neural network confidence scores during steps of differential evolution. Example of of an attack against one image, in this case the same as in Figure \ref{fig:normaltomitosis}.}
\label{fig:normaltomitosisgraph}
\end{figure}

We provide the evolutionary convergence plots for both of the example images. A convergence plot shows the change of the classifier's confidence score over the evolutionary iterations. It should be noted that the direction of the evolution depends on the type of the attack. Figure \ref{fig:mitosistonormalgraph} shows the progress of differential evolution for the adversarial image shown in Figure \ref{fig:mitosistonormal}. The lowest confidence score already reaches to almost 0.5 during the initial population attacks and drops down below 0.1 in a few steps. The minimum is reached after 40 steps. Figure \ref{fig:normaltomitosisgraph} shows the progress of differential evolution for the image and an adversarial image in Figure \ref{fig:normaltomitosis}. The maximum score of the initial population attacks is still quite low, below 0.4, but the maximum score of the population quickly rises to near 0.8 in 10 steps. The maximum 0.84 score is reached after 30 steps of the differential evolution algorithm.

\section{Discussion}

This research demonstrates that one-pixel attacks are successful against artificial neural network diagnosis of mitosis images. It shows that a machine learning model can perform acceptably with the training and testing sets but fails catastrophically when an adversarial example is used as input. In this case, the adversarial example differs by only one pixel. This hilights the need of ensuring the robustness of these artificial neural network models. While it is evident that the model works as expected in the common case, data reproduction and transmission errors, as well as cyber attacks of only one pixel, could produce undesirable results. 

It is evident that the attack against mitosis images is the easier one. These images might be of a more varied nature than the normal tissue images. Because of this, modifying the mitosis images does not create as considerable a change as when modifying normal tissue images. 
On the other hand, deceiving the artificial neural network with modified normal images was more difficult. We speculate that the neural network has likely learned to classify images with large black blobs as mitosis, thus the neural network is not easily fooled to change labels by only modifying one pixel. Larger modification of the input image would be needed for higher normal-to-mitosis attack success rate. 

This result of complete reversal of classification should not be taken as a discouragement of the use of automated diagnosis systems as part of medical imaging. Instead, it shows that the medical models built using modern artificial neural network technologies can be vulnerable to unexpected attacks and other changes in the input images. 
Attack methods will keep evolving. At the moment, the attack pixel may be somewhat prominent, which makes their detection easy, although such artifacts could be introduced just before the analysis. On the other hand, defending against such perturbations should be easy. In the future, attack-side research ideas could include blending the attack vector color values as seamlessly to the surrounding pixels as possible, thus fooling human observers. Neverhteless, detection of such attacks and robustness of image classifiers are also important research topics.

\begin{acks}
This work was supported by the Regional Council of Central Finland/Council of Tampere Region and European Regional Development Fund as part of the Health Care Cyber Range (HCCR) project of JAMK University of Applied Sciences Institute of Information Technology.

The authors would like to thank Ms.\ Tuula Kotikoski for proofreading the manuscript.
\end{acks}


\bibliographystyle{ACM-Reference-Format}
\bibliography{refs}


\begin{thebibliography}{29}


\ifx \showCODEN    \undefined \def \showCODEN     #1{\unskip}     \fi
\ifx \showDOI      \undefined \def \showDOI       #1{#1}\fi
\ifx \showISBNx    \undefined \def \showISBNx     #1{\unskip}     \fi
\ifx \showISBNxiii \undefined \def \showISBNxiii  #1{\unskip}     \fi
\ifx \showISSN     \undefined \def \showISSN      #1{\unskip}     \fi
\ifx \showLCCN     \undefined \def \showLCCN      #1{\unskip}     \fi
\ifx \shownote     \undefined \def \shownote      #1{#1}          \fi
\ifx \showarticletitle \undefined \def \showarticletitle #1{#1}   \fi
\ifx \showURL      \undefined \def \showURL       {\relax}        \fi
\providecommand\bibfield[2]{#2}
\providecommand\bibinfo[2]{#2}
\providecommand\natexlab[1]{#1}
\providecommand\showeprint[2][]{arXiv:#2}

\bibitem[\protect\citeauthoryear{Alom, Aspiras, Taha, Asari, Bowen, Billiter,
  and Arkell}{Alom et~al\mbox{.}}{2019}]%
        {alom2019advanced}
\bibfield{author}{\bibinfo{person}{Md~Zahangir Alom}, \bibinfo{person}{Theus
  Aspiras}, \bibinfo{person}{Tarek~M. Taha}, \bibinfo{person}{Vijayan~K.
  Asari}, \bibinfo{person}{T.J. Bowen}, \bibinfo{person}{Dave Billiter}, {and}
  \bibinfo{person}{Simon Arkell}.} \bibinfo{year}{2019}\natexlab{}.
\newblock \showarticletitle{Advanced deep convolutional neural network
  approaches for digital pathology image analysis: A comprehensive evaluation
  with different use cases}.
\newblock \bibinfo{journal}{\emph{arXiv preprint arXiv:1904.09075}}
  (\bibinfo{year}{2019}).
\newblock


\bibitem[\protect\citeauthoryear{Bera, Schalper, Rimm, Velcheti, and
  Madabhushi}{Bera et~al\mbox{.}}{2019}]%
        {bera2019artificial}
\bibfield{author}{\bibinfo{person}{Kaustav Bera}, \bibinfo{person}{Kurt~A.
  Schalper}, \bibinfo{person}{David~L. Rimm}, \bibinfo{person}{Vamsidhar
  Velcheti}, {and} \bibinfo{person}{Anant Madabhushi}.}
  \bibinfo{year}{2019}\natexlab{}.
\newblock \showarticletitle{Artificial intelligence in digital pathology—new
  tools for diagnosis and precision oncology}.
\newblock \bibinfo{journal}{\emph{Nature reviews Clinical oncology}}
  \bibinfo{volume}{16}, \bibinfo{number}{11} (\bibinfo{year}{2019}),
  \bibinfo{pages}{703--715}.
\newblock


\bibitem[\protect\citeauthoryear{Dusenberry and Hu}{Dusenberry and Hu}{2018}]%
        {dusenberry2018}
\bibfield{author}{\bibinfo{person}{Mike Dusenberry} {and} \bibinfo{person}{Fei
  Hu}.} \bibinfo{year}{2018}\natexlab{}.
\newblock \bibinfo{title}{Deep Learning for Breast Cancer Mitosis Detection}.
\newblock
  \bibinfo{howpublished}{https://github.com/CODAIT/deep-histopath/raw/master/docs/tupac16-paper/paper.pdf}.
\newblock


\bibitem[\protect\citeauthoryear{Feoktistov}{Feoktistov}{2006}]%
        {feoktistov2006differential}
\bibfield{author}{\bibinfo{person}{Vitaliy Feoktistov}.}
  \bibinfo{year}{2006}\natexlab{}.
\newblock \bibinfo{booktitle}{\emph{Differential evolution}}.
\newblock \bibinfo{publisher}{Springer}.
\newblock


\bibitem[\protect\citeauthoryear{Goodfellow, Shlens, and Szegedy}{Goodfellow
  et~al\mbox{.}}{2015}]%
        {goodfellow2015explaining}
\bibfield{author}{\bibinfo{person}{Ian Goodfellow}, \bibinfo{person}{Jonathon
  Shlens}, {and} \bibinfo{person}{Christian Szegedy}.}
  \bibinfo{year}{2015}\natexlab{}.
\newblock \showarticletitle{Explaining and Harnessing Adversarial Examples}. In
  \bibinfo{booktitle}{\emph{International Conference on Learning
  Representations}}.
\newblock
\urldef\tempurl%
\url{http://arxiv.org/abs/1412.6572}
\showURL{%
\tempurl}


\bibitem[\protect\citeauthoryear{{Kang}, {Song}, {Du}, and {Guizani}}{{Kang}
  et~al\mbox{.}}{2020}]%
        {Kang_2020}
\bibfield{author}{\bibinfo{person}{X. {Kang}}, \bibinfo{person}{B. {Song}},
  \bibinfo{person}{X. {Du}}, {and} \bibinfo{person}{M. {Guizani}}.}
  \bibinfo{year}{2020}\natexlab{}.
\newblock \showarticletitle{Adversarial Attacks for Image Segmentation on
  Multiple Lightweight Models}.
\newblock \bibinfo{journal}{\emph{IEEE Access}}  \bibinfo{volume}{8}
  (\bibinfo{year}{2020}), \bibinfo{pages}{31359--31370}.
\newblock
\urldef\tempurl%
\url{https://doi.org/10.1109/ACCESS.2020.2973069}
\showDOI{\tempurl}


\bibitem[\protect\citeauthoryear{Khosravi, Kazemi, Imielinski, Elemento, and
  Hajirasouliha}{Khosravi et~al\mbox{.}}{2018}]%
        {khosravi2018deep}
\bibfield{author}{\bibinfo{person}{Pegah Khosravi}, \bibinfo{person}{Ehsan
  Kazemi}, \bibinfo{person}{Marcin Imielinski}, \bibinfo{person}{Olivier
  Elemento}, {and} \bibinfo{person}{Iman Hajirasouliha}.}
  \bibinfo{year}{2018}\natexlab{}.
\newblock \showarticletitle{Deep convolutional neural networks enable
  discrimination of heterogeneous digital pathology images}.
\newblock \bibinfo{journal}{\emph{EBioMedicine}}  \bibinfo{volume}{27}
  (\bibinfo{year}{2018}), \bibinfo{pages}{317--328}.
\newblock


\bibitem[\protect\citeauthoryear{{Lin}, {Hsu}, and {Huang}}{{Lin}
  et~al\mbox{.}}{2020}]%
        {Lin_2020}
\bibfield{author}{\bibinfo{person}{B.~C. {Lin}}, \bibinfo{person}{H.~J. {Hsu}},
  {and} \bibinfo{person}{S.~K. {Huang}}.} \bibinfo{year}{2020}\natexlab{}.
\newblock \showarticletitle{Testing Convolutional Neural Network using
  Adversarial Attacks on Potential Critical Pixels}. In
  \bibinfo{booktitle}{\emph{2020 IEEE 44th Annual Computers, Software, and
  Applications Conference (COMPSAC)}}. \bibinfo{pages}{1743--1748}.
\newblock
\urldef\tempurl%
\url{https://doi.org/10.1109/COMPSAC48688.2020.000-3}
\showDOI{\tempurl}


\bibitem[\protect\citeauthoryear{{Medical Image Analysis Group Eindhoven
  (IMAG/e)}}{{Medical Image Analysis Group Eindhoven (IMAG/e)}}{2016}]%
        {tupac16}
\bibfield{author}{\bibinfo{person}{{Medical Image Analysis Group Eindhoven
  (IMAG/e)}}.} \bibinfo{year}{2016}\natexlab{}.
\newblock \bibinfo{title}{Tumor Proliferation Assessment Challenge 2016}.
\newblock \bibinfo{howpublished}{\url{http://tupac.tue-image.nl/node/3}}.
\newblock


\bibitem[\protect\citeauthoryear{Nasief, Zheng, Schott, Hall, Tsai, and amd
  X.~Allen~Li}{Nasief et~al\mbox{.}}{2019}]%
        {nasief2019machine}
\bibfield{author}{\bibinfo{person}{Haidy Nasief}, \bibinfo{person}{Cheng
  Zheng}, \bibinfo{person}{Diane Schott}, \bibinfo{person}{William Hall},
  \bibinfo{person}{Susan Tsai}, {and} \bibinfo{person}{Beth~Erickson amd
  X.~Allen~Li}.} \bibinfo{year}{2019}\natexlab{}.
\newblock \showarticletitle{A machine learning based delta-radiomics process
  for early prediction of treatment response of pancreatic cancer}.
\newblock \bibinfo{journal}{\emph{npj Precision Oncology}} \bibinfo{volume}{3},
  \bibinfo{number}{25} (\bibinfo{year}{2019}).
\newblock
\urldef\tempurl%
\url{https://doi.org/10.1038/s41698-019-0096-z}
\showDOI{\tempurl}


\bibitem[\protect\citeauthoryear{Papernot, McDaniel, Goodfellow, Jha, Celik,
  and Swami}{Papernot et~al\mbox{.}}{2017}]%
        {Papernot_2017}
\bibfield{author}{\bibinfo{person}{Nicolas Papernot}, \bibinfo{person}{Patrick
  McDaniel}, \bibinfo{person}{Ian Goodfellow}, \bibinfo{person}{Somesh Jha},
  \bibinfo{person}{Z.~Berkay Celik}, {and} \bibinfo{person}{Ananthram Swami}.}
  \bibinfo{year}{2017}\natexlab{}.
\newblock \showarticletitle{Practical Black-Box Attacks against Machine
  Learning}. In \bibinfo{booktitle}{\emph{Proceedings of the 2017 ACM on Asia
  Conference on Computer and Communications Security}} (Abu Dhabi, United Arab
  Emirates) \emph{(\bibinfo{series}{ASIA CCS '17})}.
  \bibinfo{publisher}{Association for Computing Machinery},
  \bibinfo{address}{New York, NY, USA}, \bibinfo{pages}{506–519}.
\newblock
\showISBNx{9781450349444}
\urldef\tempurl%
\url{https://doi.org/10.1145/3052973.3053009}
\showDOI{\tempurl}


\bibitem[\protect\citeauthoryear{{Paul}, {Schabath}, {Gillies}, {Hall}, and
  {Goldgof}}{{Paul} et~al\mbox{.}}{2020}]%
        {Paul_2020}
\bibfield{author}{\bibinfo{person}{R. {Paul}}, \bibinfo{person}{M. {Schabath}},
  \bibinfo{person}{R. {Gillies}}, \bibinfo{person}{L. {Hall}}, {and}
  \bibinfo{person}{D. {Goldgof}}.} \bibinfo{year}{2020}\natexlab{}.
\newblock \showarticletitle{Mitigating Adversarial Attacks on Medical Image
  Understanding Systems}. In \bibinfo{booktitle}{\emph{2020 IEEE 17th
  International Symposium on Biomedical Imaging (ISBI)}}.
  \bibinfo{pages}{1517--1521}.
\newblock
\urldef\tempurl%
\url{https://doi.org/10.1109/ISBI45749.2020.9098740}
\showDOI{\tempurl}


\bibitem[\protect\citeauthoryear{Price}{Price}{2013}]%
        {price2013differential}
\bibfield{author}{\bibinfo{person}{Kenneth~V Price}.}
  \bibinfo{year}{2013}\natexlab{}.
\newblock \showarticletitle{Differential evolution}.
\newblock In \bibinfo{booktitle}{\emph{Handbook of Optimization}}.
  \bibinfo{publisher}{Springer}, \bibinfo{pages}{187--214}.
\newblock


\bibitem[\protect\citeauthoryear{{Rajam{\"a}ki}, {Nevmerzhitskaya}, and
  {Vir{\'a}g}}{{Rajam{\"a}ki} et~al\mbox{.}}{2018}]%
        {Rajamaki2018}
\bibfield{author}{\bibinfo{person}{J. {Rajam{\"a}ki}}, \bibinfo{person}{J.
  {Nevmerzhitskaya}}, {and} \bibinfo{person}{C. {Vir{\'a}g}}.}
  \bibinfo{year}{2018}\natexlab{}.
\newblock \showarticletitle{Cybersecurity education and training in hospitals:
  Proactive resilience educational framework (Prosilience EF)}. In
  \bibinfo{booktitle}{\emph{2018 IEEE Global Engineering Education Conference
  (EDUCON)}}. \bibinfo{pages}{2042--2046}.
\newblock
\urldef\tempurl%
\url{https://doi.org/10.1109/EDUCON.2018.8363488}
\showDOI{\tempurl}


\bibitem[\protect\citeauthoryear{Schmitt}{Schmitt}{2017}]%
        {schmitt2017}
\bibfield{author}{\bibinfo{person}{Michael~N Schmitt}.}
  \bibinfo{year}{2017}\natexlab{}.
\newblock \bibinfo{booktitle}{\emph{Tallinn manual 2.0 on the international law
  applicable to cyber operations}}.
\newblock \bibinfo{publisher}{Cambridge University Press}.
\newblock


\bibitem[\protect\citeauthoryear{Sipola and Kokkonen}{Sipola and
  Kokkonen}{2021}]%
        {Sipola_2021}
\bibfield{author}{\bibinfo{person}{Tuomo Sipola} {and} \bibinfo{person}{Tero
  Kokkonen}.} \bibinfo{year}{2021}\natexlab{}.
\newblock \showarticletitle{One-Pixel Attacks Against Medical Imaging: A
  Conceptual Framework}. In \bibinfo{booktitle}{\emph{Trends and Applications
  in Information Systems and Technologies}} \emph{(\bibinfo{series}{Advances in
  Intelligent Systems and Computing}, Vol.~\bibinfo{volume}{1365})},
  \bibfield{editor}{\bibinfo{person}{{\'A}lvaro Rocha}, \bibinfo{person}{Hojjat
  Adeli}, \bibinfo{person}{Gintautas Dzemyda}, \bibinfo{person}{Fernando
  Moreira}, {and} \bibinfo{person}{Ana~Maria Ramalho~Correia}} (Eds.).
  \bibinfo{publisher}{Springer International Publishing},
  \bibinfo{address}{Cham}, \bibinfo{pages}{197--203}.
\newblock
\showISBNx{978-3-030-72657-7}
\urldef\tempurl%
\url{https://doi.org/10.1007/978-3-030-72657-7_19}
\showDOI{\tempurl}


\bibitem[\protect\citeauthoryear{Sipola, Puuska, and Kokkonen}{Sipola
  et~al\mbox{.}}{2020}]%
        {sipola2020model}
\bibfield{author}{\bibinfo{person}{Tuomo Sipola}, \bibinfo{person}{Samir
  Puuska}, {and} \bibinfo{person}{Tero Kokkonen}.}
  \bibinfo{year}{2020}\natexlab{}.
\newblock \showarticletitle{Model Fooling Attacks Against Medical Imaging: A
  Short Survey}.
\newblock \bibinfo{journal}{\emph{Information \& Security: An International
  Journal}} \bibinfo{volume}{46}, \bibinfo{number}{2} (\bibinfo{year}{2020}),
  \bibinfo{pages}{215--224}.
\newblock
\urldef\tempurl%
\url{https://doi.org/10.11610/isij.4615}
\showDOI{\tempurl}


\bibitem[\protect\citeauthoryear{{Spanakis}, {Bonomi}, {Sfakianakis},
  {Santucci}, {Lenti}, {Sorella}, {Tanasache}, {Palleschi}, {Ciccotelli},
  {Sakkalis}, and {Magalini}}{{Spanakis} et~al\mbox{.}}{2020}]%
        {Spanakis2020}
\bibfield{author}{\bibinfo{person}{E.~G. {Spanakis}}, \bibinfo{person}{S.
  {Bonomi}}, \bibinfo{person}{S. {Sfakianakis}}, \bibinfo{person}{G.
  {Santucci}}, \bibinfo{person}{S. {Lenti}}, \bibinfo{person}{M. {Sorella}},
  \bibinfo{person}{F.~D. {Tanasache}}, \bibinfo{person}{A. {Palleschi}},
  \bibinfo{person}{C. {Ciccotelli}}, \bibinfo{person}{V. {Sakkalis}}, {and}
  \bibinfo{person}{S. {Magalini}}.} \bibinfo{year}{2020}\natexlab{}.
\newblock \showarticletitle{Cyber-attacks and threats for healthcare – a
  multi-layer thread analysis}. In \bibinfo{booktitle}{\emph{2020 42nd Annual
  International Conference of the IEEE Engineering in Medicine Biology Society
  (EMBC)}}. \bibinfo{pages}{5705--5708}.
\newblock
\urldef\tempurl%
\url{https://doi.org/10.1109/EMBC44109.2020.9176698}
\showDOI{\tempurl}


\bibitem[\protect\citeauthoryear{{Stokes}, {Wang}, {Marinescu}, {Marino}, and
  {Bussone}}{{Stokes} et~al\mbox{.}}{2018}]%
        {Stokes_2018}
\bibfield{author}{\bibinfo{person}{J.~W. {Stokes}}, \bibinfo{person}{D.
  {Wang}}, \bibinfo{person}{M. {Marinescu}}, \bibinfo{person}{M. {Marino}},
  {and} \bibinfo{person}{B. {Bussone}}.} \bibinfo{year}{2018}\natexlab{}.
\newblock \showarticletitle{Attack and Defense of Dynamic Analysis-Based,
  Adversarial Neural Malware Detection Models}. In
  \bibinfo{booktitle}{\emph{MILCOM 2018 - 2018 IEEE Military Communications
  Conference (MILCOM)}}. \bibinfo{pages}{1--8}.
\newblock
\urldef\tempurl%
\url{https://doi.org/10.1109/MILCOM.2018.8599855}
\showDOI{\tempurl}


\bibitem[\protect\citeauthoryear{{Su}, {Vargas}, and {Sakurai}}{{Su}
  et~al\mbox{.}}{2019}]%
        {Su_2019}
\bibfield{author}{\bibinfo{person}{J. {Su}}, \bibinfo{person}{D.~V. {Vargas}},
  {and} \bibinfo{person}{K. {Sakurai}}.} \bibinfo{year}{2019}\natexlab{}.
\newblock \showarticletitle{One Pixel Attack for Fooling Deep Neural Networks}.
\newblock \bibinfo{journal}{\emph{IEEE Transactions on Evolutionary
  Computation}} \bibinfo{volume}{23}, \bibinfo{number}{5}
  (\bibinfo{year}{2019}), \bibinfo{pages}{828--841}.
\newblock
\urldef\tempurl%
\url{https://doi.org/10.1109/TEVC.2019.2890858}
\showDOI{\tempurl}


\bibitem[\protect\citeauthoryear{Su, Vargas, and Sakurai}{Su
  et~al\mbox{.}}{2019}]%
        {su2019one}
\bibfield{author}{\bibinfo{person}{Jiawei Su},
  \bibinfo{person}{Danilo~Vasconcellos Vargas}, {and} \bibinfo{person}{Kouichi
  Sakurai}.} \bibinfo{year}{2019}\natexlab{}.
\newblock \showarticletitle{One pixel attack for fooling deep neural networks}.
\newblock \bibinfo{journal}{\emph{IEEE Transactions on Evolutionary
  Computation}} \bibinfo{volume}{23}, \bibinfo{number}{5}
  (\bibinfo{year}{2019}), \bibinfo{pages}{828--841}.
\newblock


\bibitem[\protect\citeauthoryear{{U.S. National Cancer Institute at the
  National Institutes of Health (NIH)}}{{U.S. National Cancer Institute at the
  National Institutes of Health (NIH)}}{[n.d.]a}]%
        {NIH-1}
\bibfield{author}{\bibinfo{person}{{U.S. National Cancer Institute at the
  National Institutes of Health (NIH)}}.} \bibinfo{year}{[n.d.]}\natexlab{a}.
\newblock \bibinfo{title}{{Cancer Incidence Rate}}.
\newblock
  \bibinfo{howpublished}{\url{https://seer.cancer.gov/statistics/types/incidence.html}}.
\newblock


\bibitem[\protect\citeauthoryear{{U.S. National Cancer Institute at the
  National Institutes of Health (NIH)}}{{U.S. National Cancer Institute at the
  National Institutes of Health (NIH)}}{[n.d.]b}]%
        {NIH-2}
\bibfield{author}{\bibinfo{person}{{U.S. National Cancer Institute at the
  National Institutes of Health (NIH)}}.} \bibinfo{year}{[n.d.]}\natexlab{b}.
\newblock \bibinfo{title}{{Cancer Statistics}}.
\newblock
  \bibinfo{howpublished}{\url{https://www.cancer.gov/about-cancer/understanding/statistics}}.
\newblock


\bibitem[\protect\citeauthoryear{van Diest, van~der Wall, and Baak}{van Diest
  et~al\mbox{.}}{2004}]%
        {diest675}
\bibfield{author}{\bibinfo{person}{P~J van Diest}, \bibinfo{person}{E van~der
  Wall}, {and} \bibinfo{person}{J~P~A Baak}.} \bibinfo{year}{2004}\natexlab{}.
\newblock \showarticletitle{Prognostic value of proliferation in invasive
  breast cancer: a review}.
\newblock \bibinfo{journal}{\emph{Journal of Clinical Pathology}}
  \bibinfo{volume}{57}, \bibinfo{number}{7} (\bibinfo{year}{2004}),
  \bibinfo{pages}{675--681}.
\newblock
\showISSN{0021-9746}
\urldef\tempurl%
\url{https://doi.org/10.1136/jcp.2003.010777}
\showDOI{\tempurl}


\bibitem[\protect\citeauthoryear{Veta, Heng, Stathonikos, Bejnordi, Beca,
  Wollmann, Rohr, Shah, Wang, Rousson, Hedlund, Tellez, Ciompi, Zerhouni,
  Lanyi, Viana, Kovalev, Liauchuk, Phoulady, Qaiser, Graham, Rajpoot, Sjöblom,
  Molin, Paeng, Hwang, Park, Jia, Chang, Xu, Beck, {van Diest}, and Pluim}{Veta
  et~al\mbox{.}}{2019}]%
        {tupacpaper}
\bibfield{author}{\bibinfo{person}{Mitko Veta}, \bibinfo{person}{Yujing~J.
  Heng}, \bibinfo{person}{Nikolas Stathonikos},
  \bibinfo{person}{Babak~Ehteshami Bejnordi}, \bibinfo{person}{Francisco Beca},
  \bibinfo{person}{Thomas Wollmann}, \bibinfo{person}{Karl Rohr},
  \bibinfo{person}{Manan~A. Shah}, \bibinfo{person}{Dayong Wang},
  \bibinfo{person}{Mikael Rousson}, \bibinfo{person}{Martin Hedlund},
  \bibinfo{person}{David Tellez}, \bibinfo{person}{Francesco Ciompi},
  \bibinfo{person}{Erwan Zerhouni}, \bibinfo{person}{David Lanyi},
  \bibinfo{person}{Matheus Viana}, \bibinfo{person}{Vassili Kovalev},
  \bibinfo{person}{Vitali Liauchuk}, \bibinfo{person}{Hady~Ahmady Phoulady},
  \bibinfo{person}{Talha Qaiser}, \bibinfo{person}{Simon Graham},
  \bibinfo{person}{Nasir Rajpoot}, \bibinfo{person}{Erik Sjöblom},
  \bibinfo{person}{Jesper Molin}, \bibinfo{person}{Kyunghyun Paeng},
  \bibinfo{person}{Sangheum Hwang}, \bibinfo{person}{Sunggyun Park},
  \bibinfo{person}{Zhipeng Jia}, \bibinfo{person}{Eric I-Chao Chang},
  \bibinfo{person}{Yan Xu}, \bibinfo{person}{Andrew~H. Beck},
  \bibinfo{person}{Paul~J. {van Diest}}, {and} \bibinfo{person}{Josien~P.W.
  Pluim}.} \bibinfo{year}{2019}\natexlab{}.
\newblock \showarticletitle{{Predicting breast tumor proliferation from
  whole-slide images: The TUPAC16 challenge}}.
\newblock \bibinfo{journal}{\emph{Medical Image Analysis}}
  \bibinfo{volume}{54} (\bibinfo{year}{2019}), \bibinfo{pages}{111--121}.
\newblock
\showISSN{1361-8415}
\urldef\tempurl%
\url{https://doi.org/10.1016/j.media.2019.02.012}
\showDOI{\tempurl}


\bibitem[\protect\citeauthoryear{Veta, Van~Diest, Jiwa, Al-Janabi, and
  Pluim}{Veta et~al\mbox{.}}{2016}]%
        {veta2016mitosis}
\bibfield{author}{\bibinfo{person}{Mitko Veta}, \bibinfo{person}{Paul~J.
  Van~Diest}, \bibinfo{person}{Mehdi Jiwa}, \bibinfo{person}{Shaimaa
  Al-Janabi}, {and} \bibinfo{person}{Josien~P.W. Pluim}.}
  \bibinfo{year}{2016}\natexlab{}.
\newblock \showarticletitle{Mitosis counting in breast cancer: Object-level
  interobserver agreement and comparison to an automatic method}.
\newblock \bibinfo{journal}{\emph{PloS one}} \bibinfo{volume}{11},
  \bibinfo{number}{8} (\bibinfo{year}{2016}), \bibinfo{pages}{e0161286}.
\newblock


\bibitem[\protect\citeauthoryear{Virtanen, Gommers, Oliphant, Haberland, Reddy,
  Cournapeau, Burovski, Peterson, Weckesser, Bright, {van der Walt}, Brett,
  Wilson, Millman, Mayorov, Nelson, Jones, Kern, Larson, Carey, Polat, Feng,
  Moore, {VanderPlas}, Laxalde, Perktold, Cimrman, Henriksen, Quintero, Harris,
  Archibald, Ribeiro, Pedregosa, {van Mulbregt}, and {SciPy 1.0
  Contributors}}{Virtanen et~al\mbox{.}}{2020}]%
        {2020SciPy-NMeth}
\bibfield{author}{\bibinfo{person}{Pauli Virtanen}, \bibinfo{person}{Ralf
  Gommers}, \bibinfo{person}{Travis~E. Oliphant}, \bibinfo{person}{Matt
  Haberland}, \bibinfo{person}{Tyler Reddy}, \bibinfo{person}{David
  Cournapeau}, \bibinfo{person}{Evgeni Burovski}, \bibinfo{person}{Pearu
  Peterson}, \bibinfo{person}{Warren Weckesser}, \bibinfo{person}{Jonathan
  Bright}, \bibinfo{person}{St{\'e}fan~J. {van der Walt}},
  \bibinfo{person}{Matthew Brett}, \bibinfo{person}{Joshua Wilson},
  \bibinfo{person}{K.~Jarrod Millman}, \bibinfo{person}{Nikolay Mayorov},
  \bibinfo{person}{Andrew R.~J. Nelson}, \bibinfo{person}{Eric Jones},
  \bibinfo{person}{Robert Kern}, \bibinfo{person}{Eric Larson},
  \bibinfo{person}{C~J Carey}, \bibinfo{person}{{\.I}lhan Polat},
  \bibinfo{person}{Yu Feng}, \bibinfo{person}{Eric~W. Moore},
  \bibinfo{person}{Jake {VanderPlas}}, \bibinfo{person}{Denis Laxalde},
  \bibinfo{person}{Josef Perktold}, \bibinfo{person}{Robert Cimrman},
  \bibinfo{person}{Ian Henriksen}, \bibinfo{person}{E.~A. Quintero},
  \bibinfo{person}{Charles~R. Harris}, \bibinfo{person}{Anne~M. Archibald},
  \bibinfo{person}{Ant{\^o}nio~H. Ribeiro}, \bibinfo{person}{Fabian Pedregosa},
  \bibinfo{person}{Paul {van Mulbregt}}, {and} \bibinfo{person}{{SciPy 1.0
  Contributors}}.} \bibinfo{year}{2020}\natexlab{}.
\newblock \showarticletitle{{{SciPy} 1.0: Fundamental Algorithms for Scientific
  Computing in Python}}.
\newblock \bibinfo{journal}{\emph{Nature Methods}}  \bibinfo{volume}{17}
  (\bibinfo{year}{2020}), \bibinfo{pages}{261--272}.
\newblock
\urldef\tempurl%
\url{https://doi.org/10.1038/s41592-019-0686-2}
\showDOI{\tempurl}


\bibitem[\protect\citeauthoryear{Xu, Ma, Liu, Deb, Liu, Tang, and Jain}{Xu
  et~al\mbox{.}}{2020}]%
        {xu2020adversarial}
\bibfield{author}{\bibinfo{person}{Han Xu}, \bibinfo{person}{Yao Ma},
  \bibinfo{person}{Hao-Chen Liu}, \bibinfo{person}{Debayan Deb},
  \bibinfo{person}{Hui Liu}, \bibinfo{person}{Ji-Liang Tang}, {and}
  \bibinfo{person}{Anil~K. Jain}.} \bibinfo{year}{2020}\natexlab{}.
\newblock \showarticletitle{Adversarial attacks and defenses in images, graphs
  and text: A review}.
\newblock \bibinfo{journal}{\emph{International Journal of Automation and
  Computing}} \bibinfo{volume}{17}, \bibinfo{number}{2} (\bibinfo{year}{2020}),
  \bibinfo{pages}{151--178}.
\newblock
\urldef\tempurl%
\url{https://doi.org/10.1007/s11633-019-1211-x}
\showDOI{\tempurl}


\bibitem[\protect\citeauthoryear{Zhang, Chien, Yong, and Kuang}{Zhang
  et~al\mbox{.}}{2017}]%
        {zhang2017network}
\bibfield{author}{\bibinfo{person}{Wei Zhang}, \bibinfo{person}{Jeremy Chien},
  \bibinfo{person}{Jeongsik Yong}, {and} \bibinfo{person}{Rui Kuang}.}
  \bibinfo{year}{2017}\natexlab{}.
\newblock \showarticletitle{Network-based machine learning and graph theory
  algorithms for precision oncology}.
\newblock \bibinfo{journal}{\emph{npj Precision Oncology}} \bibinfo{volume}{1},
  \bibinfo{number}{25} (\bibinfo{year}{2017}).
\newblock
\urldef\tempurl%
\url{https://doi.org/10.1038/s41698-017-0029-7}
\showDOI{\tempurl}


\end{thebibliography}

\end{document}